\documentclass[runningheads]{llncs}
\usepackage[T1]{fontenc}

\usepackage{amsmath,amssymb,amsfonts}
\usepackage{graphicx}
\usepackage{booktabs}
\usepackage{algpseudocode}
\usepackage{hyperref}
\usepackage{caption}
\usepackage{multirow}
\usepackage{siunitx}
\usepackage{mathtools}
\usepackage[ruled,vlined]{algorithm2e}
\usepackage{subcaption}
\usepackage{cite}

\begin{document}
\title{QoS-Aware Federated Learning for Multimodal In-Cabin Interaction in Smart Vehicles\thanks{This work has been accepted and presented at WAFL workshop in ECML-PKDD 2026.}
}
\titlerunning{FedQoS for Multimodal In-Cabin Interaction in Smart Vehicles}
%
%
\authorrunning{F. Author et al.}
%
\institute{}

\author{Baran Can G{\"u}l\inst{1} \and Mert Nak{\i}p\inst{2, 3} \and
Nasser Jazdi\inst{1} \and
Michael Weyrich\inst{1}}
\authorrunning{B. C. G{\"u}l et al.}
%
\institute{Institute of Industrial Automation and Software Engineering, University of Stuttgart,
Pfaffenwaldring 47, 70550 Stuttgart, Germany\\
\email{\{baran-can.guel, nasser.jazdi, michael.weyrich\}@ias.uni-stuttgart.de}
\and
Institute of Theoretical and Applied Informatics, Polish Academy of Sciences (IITIS PAN), 44-100 Gliwice, Poland\\
\email{mnakip@iitis.pl}\\
\and
Ideatrum Ltd. Sti.,  35560, İzmir, Türkiye\\}

\maketitle  

\begin{abstract}
Modern smart vehicles leverage multimodal sensors, ranging from high-bandwidth vision systems to low-rate physiological monitors, to provide personalized in-cabin services. However, integrating high-fidelity multimodal fusion with collaborative training is often hindered by the heterogeneous and time-varying Quality of Service (QoS) constraints of vehicular networks. Standard Federated Learning (FL) approaches enforce rigid synchronous rounds that fail to account for these resource asymmetries, leading to safety-critical timing violations and energy exhaustion. In this paper, we propose FedQoS, a novel asynchronous, event-triggered FL framework that decouples local computation from global communication via a two-phase gating mechanism. First, we introduce a resource-aware training gate that initializes local learning only when sensing buffers and energy reserves meet safety thresholds, preventing ML tasks from compromising core vehicle mobility. Second, a QoS-aware transmission policy gates uplink updates based on an efficiency score that balances model novelty against instantaneous latency and energy costs. Locally, clients optimize an objective featuring a staleness-aware proximal term that dynamically adjusts the global anchor strength based on update age. Extensive experiments on multimodal vehicular datasets demonstrate that FedQoS achieves competitive personalized accuracy with only marginal performance loss compared to FedAvg, while substantially reducing QoS violations, cutting communication overhead by 76.7\%, and lowering latency cost by 26.0\%, demonstrating a highly favorable accuracy and efficiency balance for real-world vehicular deployments.
\end{abstract}

\section{Introduction}

Smart vehicles are increasingly equipped with multimodal sensors capable of monitoring driver and passenger states as well as ambient conditions~\cite{Guel24}. These sensors, ranging from environmental monitors (e.g., temperature, humidity) to camera-based facial scanners and physiological signal detectors (e.g., heart rate, skin conductance), generate rich data streams that can be leveraged to enhance in-cabin safety, comfort, and overall user experience~\cite{kim2022multimodal, bala2020driver, Guel24_2}. However, these modalities are inherently asymmetric; for instance, computer-vision-based drowsiness detection is significantly more resource-intensive than low-bandwidth physiological monitoring~\cite{bala2020driver}. In dynamic vehicular environments, the ability of the system to process these heterogeneous data streams fluctuates according to strict Quality of Service (QoS) boundaries.

While Federated Learning (FL) has emerged as a privacy-preserving paradigm for distributed machine learning~\cite{mcmahan2017communication}, existing approaches face a fundamental disconnect when deployed in vehicular environments: \emph{they optimize for model convergence without considering the heterogeneous, time-varying resource constraints that govern real-world vehicle operations}. This creates three critical failures:
\begin{enumerate}
    
    \item \textbf{Safety-Critical Timing Violations:} Standard FL methods fail to prioritize heavy modalities. While drowsiness detection requires inference latency $<$100\,ms~\cite{bala2020driver}, methods such as FedAvg~\cite{mcmahan2017communication} assign uniform computational budgets, incurring an average inference latency cost of 1.598\,s, exceeding the safety threshold by more than an order of magnitude in resource-constrained vehicular ECUs (see Section~\ref{sec:results}).
    
    \item \textbf{Energy Budget Exhaustion:} High-fidelity multimodal fusion increases power draw in electric vehicles, where ML inference competes directly with propulsion~\cite{aloqaily2022energy}. Existing FL frameworks ignore energy/accuracy trade-offs. Such that, under FedAvg, clients fully exhaust their onboard energy budget during training (see Section~\ref{sec:results}), reducing the margin for core mobility functions.
    
    
    \item \textbf{Personalization-Resource Conflict:} Current personalized FL methods (FedPer~\cite{fedper}, APFL~\cite{apfl}) improve per-client accuracy but \emph{increase} computational complexity, directly contradicting QoS requirements in resource-limited settings.
\end{enumerate}

We recognize that QoS constraints and personalization objectives are not orthogonal challenges but rather coupled optimization problems that must be solved jointly. A vehicle with limited battery should not only train a personalized model but should do so with an energy-aware objective that trades off accuracy for battery preservation. Similarly, safety-critical predictions should trigger dynamic resource allocation that prioritizes latency over model complexity.

This paper introduces FedQoS, a novel event-triggered Federated Learning (FL) framework specifically designed for the resource-constrained and asynchronous nature of vehicular networks. Unlike prior work that treats Quality of Service (QoS) as a static penalty or post-hoc filter~\cite{hudson2021qos, hierarchialflqos2022}, FedQoS governs the learning lifecycle through a distinct two-phase gating mechanism. In the first phase, local training is only initialized if sensing buffers and residual energy levels meet safety-critical thresholds. When triggered, the vehicle optimizes a local objective featuring a time-decaying proximal term that accounts for model staleness:
\begin{equation}
\min_{\theta_i} \; \underbrace{\mathcal{C}_{\text{task}}(\theta_i; D_i^{(t_k)})}_{\text{Local Task Accuracy}} + \underbrace{\mu_i(t_k) \|\theta_i - \theta_g\|_2^2}_{\text{Staleness-Aware Personalization}}
\end{equation}
where $\mu_i(t_k)$ is a dynamically adjusted regularization coefficient that exponentially decays based on the time elapsed since the last global update. This formulation ensures that vehicles maintain consistency with the global model when connectivity is fresh but are permitted to prioritize local personalization as the global prior becomes obsolete.

Following local optimization, the second phase employs a \emph{transmission efficiency score} to determine whether to offload the update or cache it, balancing information utility against real-time latency and energy costs. By decoupling computation from communication, FedQoS enables three unique capabilities: 
(1) \emph{Resource-reactive training}, ensuring ML tasks do not compromise core mobility functions; (2) \emph{Staleness-aware personalization}, which prevents outdated global models from degrading local performance; and (3) \emph{Cost-optimized offloading}, which triggers transmissions only under favorable channel and battery conditions. 


Extensive empirical validation on a real-world multimodal vehicular dataset demonstrates that this event-triggered approach achieves competitive personalized accuracy with only marginal performance loss relative to FedAvg, while reducing total communication volume by 76.7\%, eliminating wasted transmissions entirely, lowering average latency cost by 26.0\% (from 1.598\,s to 1.182\,s), and reducing energy consumption by 10.8\%, establishing a strongly favorable accuracy and efficiency balance for safety-critical vehicular deployments.

The rest of the paper is organized as follows: Section~\ref{sec:related_work} structurally reviews the recent related literature. Section~\ref{sec:methodology} presents the methodology of the FedQoS framework. Section~\ref{sec:results} presents the performance evaluation results on a multimodal in-vehicle dataset..  Section~\ref{sec:conc} provides the summary of this work and insight for future research directions.

\section{Related Work}\label{sec:related_work}
Federated Learning (FL) has emerged as a promising distributed machine learning paradigm that addresses the limitations of centralized learning, such as high transmission overhead, privacy concerns, and reliance on single points of failure. This section reviews relevant literature on FL in Vehicular Networks and QoS-aware FL, highlighting contributions and identifying current research gaps.

The application of FL in vehicular networks has garnered significant attention for enabling collaborative intelligence while preserving data privacy in dynamic and resource-constrained environments, with early work highlighting critical challenges including data heterogeneity, stringent latency requirements, bandwidth constraints, and security concerns \cite{elbir2022federated, nakip2023decentralized, gelenbe2024disfida}. To address scalability, Reference~\cite{posner2021federated} proposed a Federated Vehicular Cloud (FVC) architecture leveraging heterogeneous communication and Blockchain systems, while Reference~\cite{quan2025federated} integrated hybrid multi-agent deep reinforcement learning with FL to improve task completion and reduce delay and energy consumption in IoV systems. Reference~\cite{zhang2024vehicle} further developed a mobility- and channel-aware FL system enabling RSUs to perform weighted model aggregation under dynamic conditions, and Reference~\cite{aloqaily2022energy} strengthened FL by integrating UAVs and UGVs with a blockchain-supported framework to ensure continuous connectivity and data integrity.

Privacy preservation and personalization have been equally central concerns, with works addressing noise-based privacy mechanisms \cite{qu2023privacy}, blockchain-secured location prediction \cite{ali2025federated}, and data heterogeneity through personalized FL approaches \cite{liao2025personalized, zheng2022data, wu2025personalized, prathiba2021cybertwin}. On the QoS front, only a limited number of works have integrated QoS considerations into FL systems \cite{huang2023federated, tam2023enhancing, abdulrahman2023, zhou2024enhancing}, including QoS-aware caching in D2D fog networks \cite{huang2023federated}, congestion-aware edge aggregation scheduling \cite{tam2023enhancing}, clustering-based overhead reduction for vehicular FL \cite{abdulrahman2023}, and edge-cloud QoS optimization for smart transportation \cite{zhou2024enhancing}.

Despite this progress, existing approaches exhibit three fundamental limitations. First, personalized FL methods treat system resources as infinite while QoS-aware methods filter clients without adapting the learning lifecycle \cite{qosfl2023}; we address this via an event-triggered gating mechanism coupling local training initialization to real-time vehicle energy and buffer states. Second, prior QoS-aware FL \cite{zhang2024vehicle, oraclefed, syncfed} relies on fixed thresholds that ignore vehicular dynamism and model staleness; FedQoS resolves this through a staleness-aware personalization objective with a time-decaying coefficient $\mu_i(t_k)$ that dynamically balances global consistency and local adaptation. Third, existing FL work ignores heterogeneous sensing modalities and resource-reactive training; FedQoS addresses this via a two-phase gating protocol ensuring high-fidelity multimodal fusion is only executed when the vehicle's QoS profile permits.

\begin{figure}[h!]
    \centering
    \includegraphics[width=0.75\textwidth]{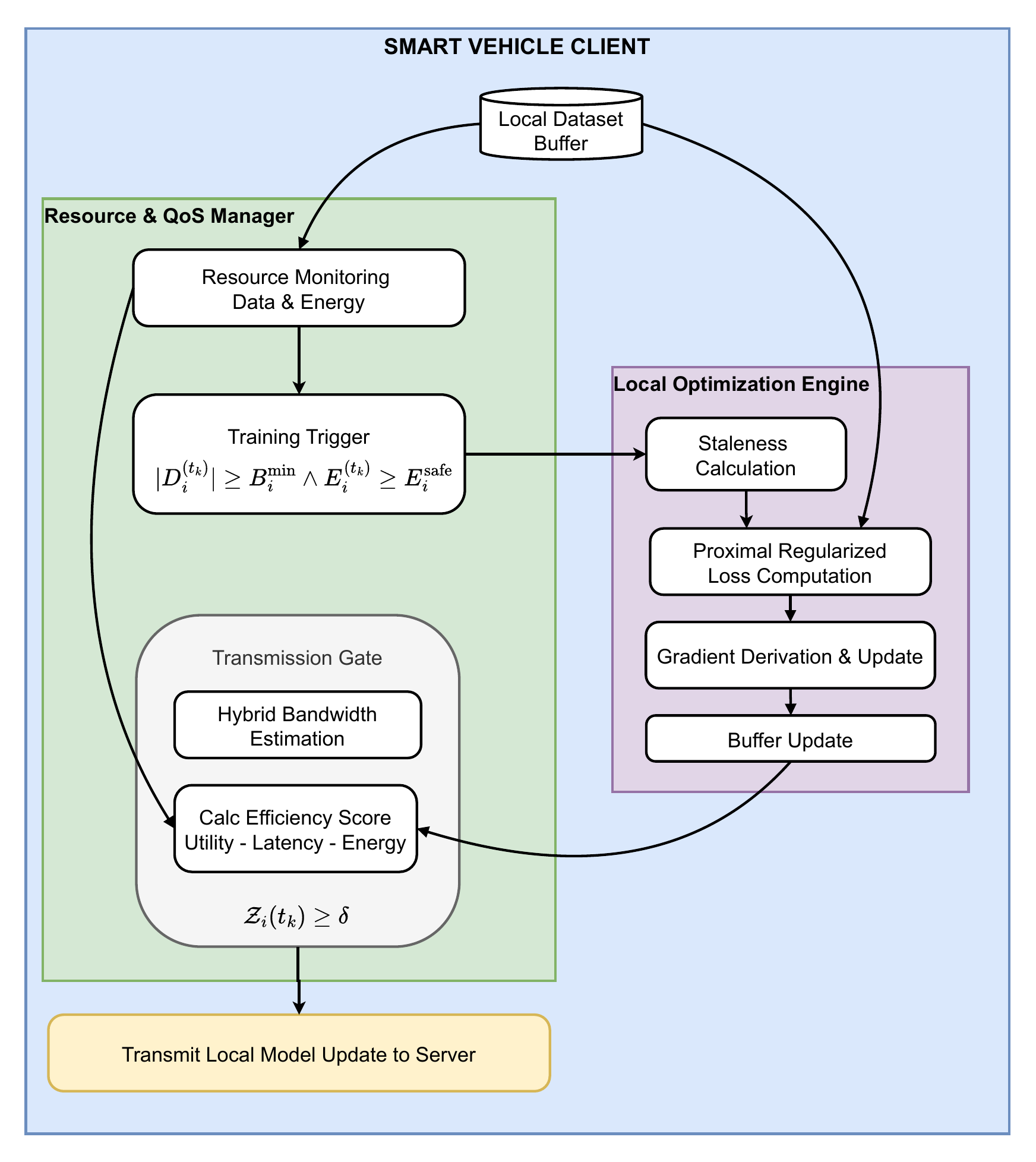}
    \caption{Overview of FedQoS: an event-triggered, QoS-aware federated learning framework for personalized multimodal in-cabin driver state monitoring. 
    }
    \label{fig:placeholder}
\end{figure}

\section{QoS-Aware Federated Architecture with Event-Triggered Update Scheduling}\label{sec:methodology}

In this section, we propose an asynchronous, event-driven framework that decouples local computation from global communication. Unlike standard Federated Learning, which enforces synchronous rounds with fixed frequency, our approach governs the learning and transmission processes via a distinct Two-Phase Gating Mechanism. This ensures that computational resources are expended only when sufficient data and energy are available, and that communication occurs only when the Quality of Service (QoS) conditions allow for efficient offloading.

To this end, we consider a federated system comprising $N$ smart vehicles $\mathcal{V} = \{v_1, \ldots, v_N\}$ and a central server. This two-tier architecture reflects the standard asynchronous federated learning architecture where clients perform local training and the server coordinates global model updates through iterative aggregation. The system operates asynchronously, with each vehicle independently triggering local events based on its sensing and resource state.

\subsection{Resource-Aware Training Phase}
The first phase governs the initialization of local training. This process is designed to be reactive to the sensing environment of the vehicle while protecting its energy reserves. To this end, in this phase, FedQoS continuously monitors the number of samples for training and the current battery level. 

Let $D_i^{(t_k)}$ denote the buffer of fresh sensing data accumulated by vehicle $i$ during the interval $(t_{k-1}, t_k]$, where $t_{k-1}$ marks the completion of the previous training round. This accumulated batch of data $D_i^{(t_k)} = \{(x_{i,j}, y_{i,j})\}_{j=1}^{|D_i^{(t_k)}|}$, where $x_{i,j}$ represents the input features (e.g., sensor readings, camera frames) and $y_{i,j}$ represents the corresponding ground truth values for the $j$-th sample. Also, let $E_i^{(t_k)}$ denote the residual energy budget available on vehicle $i$ at the triggering instance $t_k$. This instantaneous state $E_i^{(t_k)}$ represents the remaining battery capacity after deducting the cumulative energy consumption incurred by safety-critical mobility and continuous sensing operations during the interval $(t_{k-1}, t_k]$.

A local training event is triggered at time $t_k$ if and only if the accumulated sample size satisfies the minimum batch requirement and the residual energy exceeds the safety threshold. Accordingly, we define the binary trigger indicator for training, denoted by $I_{\text{train}}^{(t_k)}$ as:
\begin{equation}
\label{eq:training_trigger}
I_{\text{train}}^{(t_k)} = 
\begin{cases} 
1 & \text{if } |D_i^{(t_k)}| \ge B_i^{\min} \land E_i^{(t_k)} \ge E_i^{\text{safe}}, \\
0 & \text{otherwise},
\end{cases}
\end{equation}
where $B_i^{\min}$ represents the minimum batch size required for stable stochastic gradient estimation, and $E_i^{\text{safe}}$ denotes the critical energy threshold reserved to ensure that the core safety and mobility functions of vehicle $i$ are not compromised by the learning process.

Upon triggering ($I_{\text{train}}=1$), the vehicle executes local training. Unlike synchronous FL where the global model $\theta_g$ is always fresh, in our asynchronous setting, the local $\theta_g$ may be stale. Let $\tau_i$ denote the timestamp when the current global model $\theta_g$ was downloaded.

In order to account for this, we formulate the local objective with a Time-Decaying Proximal Term:
\begin{equation}
\label{eq:local_objective}
\min_{\theta_i} \; \Big( \mathcal{C}_{\text{task}}(\theta_i; D_i^{(t_k)}) + \mu_i(t_k) \|\theta_i - \theta_g\|_2^2 \Big),
\end{equation}
where, $\mathcal{C}_{\text{task}}$ is the empirical task loss (e.g., categorical cross-entropy). The regularization coefficient $\mu_i(t_k)$ controls the strength of the global anchor. 

We define $\mu_i(t_k)$ as an exponentially decaying function of the staleness duration $(t_k - \tau_i)$:
\begin{equation}
\mu_i(t_k) = \mu_0 \, e^{\left( -\alpha (t_k - \tau_i) \right)},
\end{equation}
where $\mu_0$ is the nominal regularization strength and $\alpha > 0$ is a decay rate. During the learning process, if the global model is fresh ($t_k \approx \tau_i$), $\mu_i \approx \mu_0$, enforcing strong consistency with the server. On the other hand, if the global model is obsolete ($t_k \gg \tau_i$), $\mu_i \to 0$, effectively decoupling the client. This allows the vehicle to continue learning from its local data stream without being constrained by an outdated global prior.

\subsection{QoS-Aware Transmission Gating}
Once a local update $\theta_i^{(t_k)}$ is computed, the system enters the second phase to determine whether to transmit the update or cache it. This decision is modeled as an instantaneous multi-objective optimization problem. To this end, we define the Transmission Efficiency Score, $\mathcal{Z}_i(t_k)$, which balances the utility of the update against the communication costs:
\begin{equation}
\mathcal{Z}_i(t_k) = w_m \mathcal{U}(t_k) - w_l \mathcal{L}(t_k) - w_e \mathcal{E}(t_k).
\end{equation}

The final transmission decision $a_i(t_k) \in \{0, 1\}$ is taken as follows:
\begin{equation}
a_i(t_k) = 
\begin{cases} 
1 \quad (\text{Transmit}) & \text{if } \mathcal{Z}_i(t_k) \ge \delta, \\
0 \quad (\text{Buffer}) & \text{otherwise}.
\end{cases}
\end{equation}
If $a_i=0$, the update is buffered locally. The vehicle periodically re-evaluates the transmission decision as channel conditions or battery states evolve, eventually releasing the update when the cost becomes acceptable. During this period, the vehicle may decide to perform new training processes.

\subsubsection{Information Utility -- $\mathcal{U}(t_k)$}
With the trained model $\theta_i^{(t_k)}$ now available, we quantify its novelty by calculating the Euclidean divergence from the global reference:
\begin{equation}
\mathcal{U}(t_k) = \|\theta_i^{(t_k)} - \theta_g\|_2^2.
\end{equation}
A high divergence suggests that the vehicle has learned significant new features not yet captured by the global model, thereby increasing the priority of this update.

\subsubsection{Latency Cost -- $\mathcal{L}(t_k)$}
This term penalizes transmission under poor network conditions. We estimate the latency cost $\mathcal{L}(t_k)$ based on the current uplink bandwidth $b_i^{(t_k)}$ and channel reliability $r_i$:
\begin{equation}
\mathcal{L}(t_k) = \frac{\|\theta_i^{(t_k)}\|_1}{b_i^{(t_k)} \cdot r_i},
\end{equation}
where $\|\theta_i^{(t_k)}\|_1$ approximates the compressed payload size. The bandwidth $b_i^{(t_k)}$ is obtained via the hybrid passive-active estimation.

First, we utilize a passive estimator based on historical transmission performance. Let $V_{\text{last}}$ be the volume (in bits) of the last successfully transmitted payload and $\Delta t_{\text{last}}$ be the duration of that transmission event. The instantaneous bandwidth sample is calculated as $b_{\text{sample}} = V_{\text{last}} / \Delta t_{\text{last}}$, and smoothed via an Exponential Moving Average (EMA) with factor $\nu \in (0,1)$.

However, to mitigate staleness during long idle intervals, if the time elapsed since the last update exceeds a coherence threshold $T_{\text{coh}}$, the vehicle triggers a lightweight active probe using periodic safety beacons. The bandwidth estimate is then refreshed using the Signal-to-Noise Ratio (SNR) of these pilot signals:
\begin{equation}
b_i^{(t_k)} = 
\begin{cases} 
\nu \cdot b_{\text{sample}} + (1 - \nu) \cdot b_i^{(t_{k-1})}, & \text{if } t_{\text{idle}} \le T_{\text{coh}} \\
W \cdot \log_2(1 + \text{SNR}_{\text{pilot}}), & \text{otherwise}.
\end{cases}
\end{equation}
This ensures the system does not make offloading decisions based on obsolete channel data.

\subsubsection{Energy Impact -- $\mathcal{E}(t_k)$}
We penalize transmission when the battery is low relative to the required transmission effort. The energy impact ratio is defined as:
\begin{equation}
\mathcal{E}(t_k) = \frac{P_{\text{tx}} \, \mathcal{L}(t_k)}{E_i^{(t_k)}},
\end{equation}
where $P_{\text{tx}}$ is the transmission power and $E_i^{(t_k)}$ is the residual energy budget. This term grows asymptotically as the battery depletes ($E_i \to 0$), effectively blocking non-critical updates on energy-starved devices.

\subsection{Server Aggregation}
The server operates in an asynchronous manner, collecting model updates $\Delta\theta_i$ from the subset of clients $\mathcal{S}_t$ that successfully triggered the offloading condition ($a_i=1$) within the current aggregation window. Note that due to the event-triggered nature of the system, updates may arrive with varying degrees of staleness. Let $\tau_i$ denote the timestamp of the global model version used by client $i$ to compute its update.

The server gathers the updates and aggregates them using staleness-aware weights:
\begin{equation}
S_i^{(t)} = \frac{w_i \, \phi(r_i) \, \rho(t - \tau_i)}{\sum_{j \in \mathcal{S}_t} w_j \, \phi(r_j) \, \rho(t - \tau_j)},
\end{equation}
where $w_i = |D_i^{(t_k)}|$ is the batch size, and $\phi(r_i)$ accounts for link reliability. The function $\rho(\Delta t) = (1 + \Delta t)^{-\beta}$ is a staleness dampening function (with $\beta > 0$) that reduces the impact of updates computed on obsolete global hypotheses, thereby stabilizing the asynchronous convergence. The global model is then updated as:
\begin{equation}
\theta_g \leftarrow (1-\eta_g)\theta_g + \eta_g \sum_{i\in\mathcal{S}_t} S_i^{(t)} \Delta\theta_i,
\end{equation}
where $\eta_g \in (0,1]$ is a global mixing parameter. The updated global reference $\theta_g$ is subsequently broadcast to the network, serving as the new anchor point for the next cycle of event-triggered local learning.

\section{Experimental Results}
\label{sec:results}

This section evaluates the effectiveness of FedQoS in multimodal vehicular learning scenarios under heterogeneous system constraints. We compare FedQoS against FedAvg across two complementary dimensions: (i)~personalized predictive performance over communication rounds, and (ii)~QoS efficiency under realistic vehicular network conditions.

\subsection{Experimental Setup}
Experiments are conducted on a real-world multimodal driver state monitoring dataset comprising synchronized vehicle telemetry, physiological signals (PPG and EEG), and cabin-facing video. Data are collected from multiple participants across six driving conditions: \emph{normal}, \emph{distracted}, \emph{alerted}, \emph{stressed}, \emph{relaxed}, and \emph{drowsy}. Each driving session is segmented into fixed-length temporal windows to enable sequential modeling. Data are partitioned into training and test sets using subject-level temporal splits.

Our federated learning system is simulated with $N{=}10$ clients over 20 rounds on the real-world dataset, representing heterogeneous vehicles with non-IID data distributions. Each client performs three local epochs before transmitting updates. The server aggregates received updates using staleness-aware weighting with decay factor $\beta_{\text{stale}}{=}0.5$ and global learning rate $\alpha_{\text{global}}{=}0.1$.

QoS modeling contains time-varying network conditions that are simulated by varying bandwidth, latency, and channel reliability within realistic vehicular ranges (total simulation window: 300\,s). Client energy budgets are finite ($B_{\text{total}}{=}1000$\,J per client), and communication costs depend on both model payload size and instantaneous network conditions. QoS parameters drive FedQoS's two-phase gating mechanism: updates that fail either the QoS gate or the energy gate are suppressed before transmission.

\subsection{Multimodal Architecture}

We employ modality-specific encoders aligned with the characteristics of each data source: a two-layer stacked BiLSTM (64 and 32 hidden units per direction) for vehicle telemetry (14 features) to capture bidirectional temporal dependencies; a two-layer LSTM (128 hidden units) with early feature-level fusion for physiological signals (PPG and EEG, 7 features) to model correlated temporal dynamics; and a four-layer 3D-CNN (3$\to$32$\to$64$\to$128$\to$128 channels) for cabin-facing facial video to extract spatiotemporal representations. Each encoder projects its output to a shared 256-dimensional embedding space. The resulting modality embeddings are combined using an 8-head cross-modal self-attention fusion module with a learnable modality gate, projecting to a 512-dimensional fused representation, followed by a two-layer MLP classifier head (512$\to$256$\to$6 classes). Model architectures are fixed across all methods to ensure fair comparison.

\subsection{Performance Evaluation}
\label{sec:results}
\begin{figure}[h!]
  \centering
  \includegraphics[width=0.95\linewidth]{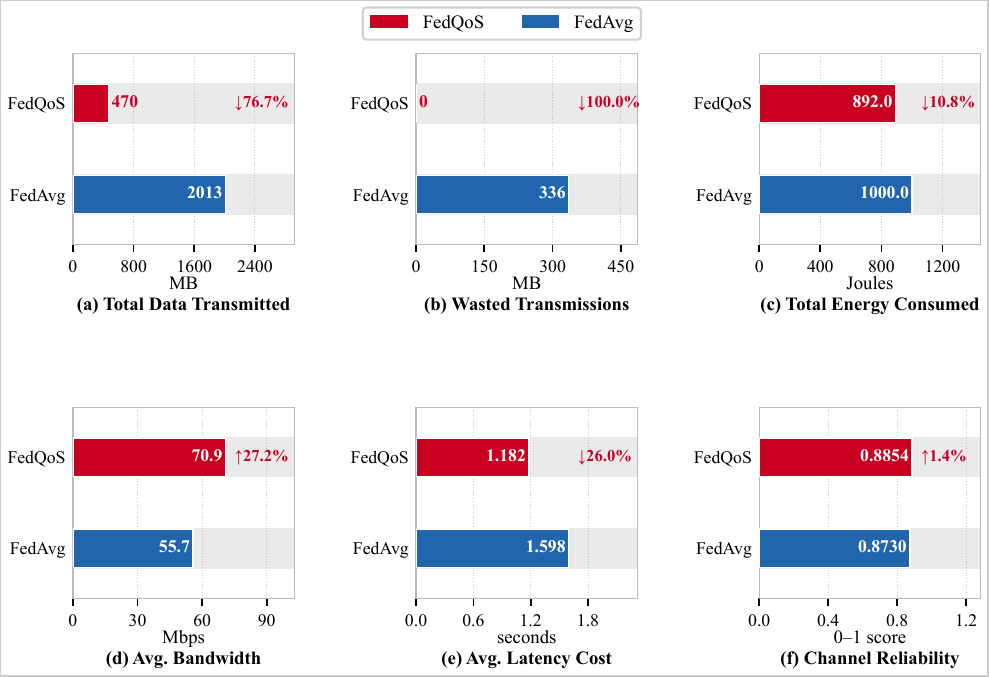}
  \caption{QoS efficiency comparison between FedQoS and FedAvg across six communication metrics. FedQoS reduces total data transmission by 76.7\%, eliminates wasted transmissions entirely via its gating mechanism, lowers energy consumption by 10.8\%, and achieves higher average bandwidth and lower latency cost compared to FedAvg.}
  \label{fig:qos_advantages}
\end{figure}

We now present the performance evaluation of the proposed FedQos framework. To this end, first, Figure~\ref{fig:qos_advantages} compares FedQoS and FedAvg across six QoS efficiency metrics measured during the full 300\,s simulation under realistic vehicular network conditions. These six metrics are analyzed in three main categories: Communication Volume (Figure~\ref{fig:qos_advantages}(a)-(b)), Energy Consumption (Figure~\ref{fig:qos_advantages}(c)), and Bandwidth and Latency (Figure~\ref{fig:qos_advantages}(d)-(f)).

FedQoS transmits only $469.7$\,MB of model updates over the course of training, compared to $2013.1$\,MB for FedAvg, a \textbf{76.7\% reduction} in total communication volume (see Figure~\ref{fig:qos_advantages}(a)). This saving is achieved by the two-phase gating mechanism, which blocks transmissions that fail the QoS threshold ($85.3\%$ QoS block rate) or exceed the remaining energy budget of the client. Crucially, all $0.0$\,MB of FedQoS's blocked updates are discarded before transmission, meaning \emph{zero} bandwidth is wasted on updates that would never contribute to a successful aggregation. By contrast, FedAvg wastes $335.5$\,MB on transmissions that are subsequently lost due to channel failures, incurring both energy and latency costs with no learning benefit as shown in Figure~\ref{fig:qos_advantages}(b). 

\vspace{-4mm}

\vspace{-4mm}

The results in Figure~\ref{fig:qos_advantages}(c) reveal that FedQoS consumes $892.1$\,J in total, representing a \textbf{10.8\% energy saving} over FedAvg's $1000.0$\,J budget. While the absolute saving appears moderate, it is significant in the vehicular context where on-board compute and battery budgets are strictly bounded. The energy gate in FedQoS ensures that clients approaching their budget ceiling contribute to the global model only when doing so is energetically viable, preventing premature client dropout mid-training.

Figure~\ref{fig:qos_advantages}(d) and (e) show that FedQoS achieves an average bandwidth of $70.9$\,Mbps compared to $55.7$\,Mbps for FedAvg (\textbf{+27.2\%}), and reduces average latency cost from $1.598$\,s to $1.182$\,s (\textbf{$\downarrow$26.0\%}). These improvements stem from the QoS-aware client selection policy: by preferentially including clients with high instantaneous channel quality and low latency, FedQoS ensures that accepted transmissions traverse better network conditions than the indiscriminate inclusion strategy of FedAvg. Average channel reliability improves correspondingly from $0.8730$ to $0.8854$ as given in Figure~\ref{fig:qos_advantages} (f).

\begin{figure}[h!]
    \centering
    \begin{subfigure}[b]{0.49\textwidth}
        \centering
        \includegraphics[width=\linewidth]{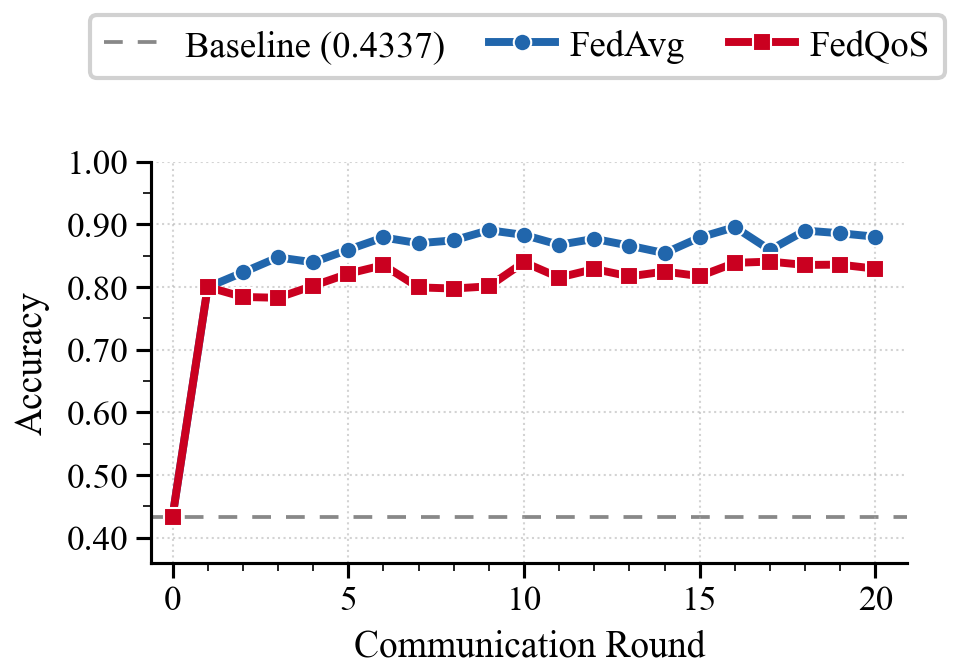}
        \caption{Average test accuracy}
        \label{fig:acc}
    \end{subfigure}
    \hfill
    \begin{subfigure}[b]{0.49\textwidth}
        \centering
        \includegraphics[width=\linewidth]{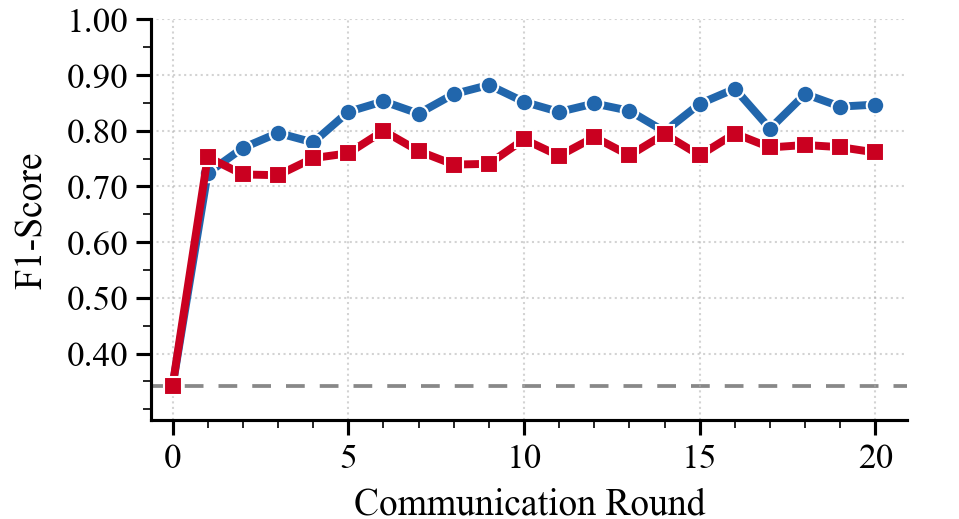}
        \caption{Average F1-score}
        \label{fig:f1}
    \end{subfigure}

    \caption{Performance of FedAvg and FedQoS over 20 communication rounds. Both methods share the same pre-training baseline (dashed line). Results are averaged across all 10 participating clients.}
    \label{fig:acc_f1}
\end{figure}

Furthermore, Figure~\ref{fig:acc_f1} illustrates the average client test accuracy and F1-score of \textit{FedAvg} and \textit{FedQoS} over 20 communication rounds. Both methods are initialized from the same pre-training baseline (accuracy: 0.4337, F1: 0.3415). FedAvg converges rapidly within the first five rounds and stabilizes around 88--90\% accuracy, while FedQoS achieves comparable accuracy with the additional benefit of QoS-aware client selection, as reflected by the communication efficiency gains.

\vspace{-4mm}

\vspace{-4mm}
The results, displayed in Figure~\ref{fig:qos_advantages} and Figure~\ref{fig:acc_f1}, reveal a clear and quantifiable accuracy/efficiency trade-off. FedAvg maximizes predictive performance at the cost of $4.3\times$ more transmitted data, $335.5$\,MB of wasted bandwidth, and higher per-round latency. FedQoS achieves $95\%$ of the accuracy that FedAvg achieves with dramatically lower communication overhead, zero wasted transmissions, and substantially better utilization of available channel capacity.
In bandwidth-constrained vehicular deployments, where erratic connectivity and limited energy reserves are first-class constraints rather than afterthoughts, this trade-off strongly favors FedQoS. 

Overall, the experimental results show that FedQoS effectively balances personalization, global coordination, and system constraints. By coupling QoS-aware regularization with asynchronous aggregation, FedQoS achieves strong predictive performance while substantially reducing communication and energy costs, making it well-suited for real-world vehicular federated learning.

\section{Conclusion and Future Work}
\label{sec:conc}
This paper presented \textbf{FedQoS}, an asynchronous, event-triggered federated learning framework designed to reconcile the pursuit of high-performance personalization with the volatile resource constraints of modern smart vehicles. By decoupling local computation from global communication through a two-phase gating mechanism, the framework ensures that resource-intensive multimodal fusion is only executed when energy reserves and data buffers meet safety-critical thresholds. Furthermore, by introducing a staleness-aware proximal objective, FedQoS allows vehicles to dynamically adjust their consistency with the global model based on the age of received parameters, facilitating stable learning even under intermittent connectivity.

Empirically, FedQoS directly addresses the two critical failure modes of conventional FL: the safety-critical timing violation and the energy budget exhaustion incurred by methods such as FedAvg in resource-constrained vehicular ECUs. Across all evaluated QoS dimensions, FedQoS delivers superior system-level efficiency. It reduces total communication volume by 76.7\%, eliminates wasted transmissions entirely via its gating mechanism, lowers average latency cost by 26.0\%, and improves channel utilization, all while incurring only a marginal accuracy trade-off of approximately 5\% relative to FedAvg. Crucially, this performance loss is acceptable in vehicular deployments, where QoS reliability and energy sustainability are first-class constraints. One may say that a model that converges reliably under strict energy and latency budgets is strictly preferable to one that achieves marginally higher accuracy at the cost of safety-critical deadline violations and premature battery depletion.

In summary, FedQoS achieves considerably better QoS performance with a highly competitive accuracy compared to FedAvg, making it a practical and scalable solution for personalized multimodal learning in heterogeneous automotive platforms.

Future research will explore per-modality online learning strategies, enabling the framework to dynamically prioritize and update modality-specific sub-networks based on instantaneous data importance and available hardware resources. We also aim to extend our experimental validation to realistic networking simulations incorporating high-fidelity vehicular mobility models and complex signal propagation environments, further refining FedQoS as a robust solution for the evolving landscape of multimodal intelligent transportation systems.

\section*{Acknowledgment}
This publication is carried out as part of the Hardware Abstraction Layer for Software Defined Vehicle (HAL4SDV) Project No. 101139789, which is co-funded by the European Union (Chips JU).

A part of research leading to these results is in the frame of the PACK - Persistent Awareness system facilitating Customs checKs - project, which has received funding from the European Union's Horizon Europe research and innovation programme under the Grant Agreement No. 101225875. The views and opinions expressed in this document are the sole responsibility of the authors and do not necessarily reflect the views or positions of the European Commission. Neither the European Union nor the granting authority can be held responsible for them.

In addition, Mert Nak{\i}p gratefully acknowledges that his work is supported by the Foundation for Polish Science (FNP) under agreement no. START 057.2025. 

\bibliographystyle{plain}
\bibliography{bibtex/bib/references,bibtex/bib/ref_relatedworks}

\end{document}